\documentclass[letterpaper, 10 pt, conference]{ieeeconf}  

\IEEEoverridecommandlockouts                              
\let\labelindent\relax
\usepackage[loadonly]{enumitem} 
\usepackage[utf8]{inputenc} 
\usepackage{graphicx}
\usepackage{amsmath}         
\usepackage{amssymb}         
\usepackage{textcomp} 
\usepackage{booktabs}
\usepackage{cite}  
\usepackage{colortbl}
\usepackage{textcomp}
\usepackage{multicol}
\usepackage{siunitx}
\usepackage{tabularx,booktabs}              
\usepackage{caption}
\usepackage{makecell}   
\usepackage{adjustbox}  
\usepackage{cuted}  
\usepackage{hyperref}

\usepackage{multirow,siunitx}
\usepackage[table]{xcolor}   
\newcolumntype{L}{>{\raggedright\arraybackslash}X} 

\title{\LARGE \bf
RT-Cache: Training-Free Retrieval for Real-Time Manipulation
}

\author{Owen Kwon$^{1}$, Abraham George$^{2}$, Alison Bartsch$^{2}$, and Amir Barati Farimani$^{1,2}$%
  \thanks{$^{1}$Department of Biomedical Engineering, Carnegie Mellon University, Pittsburgh PA 15213, USA. 
  }
  \thanks{$^{2}$Department of Mechanical Engineering, Carnegie Mellon University, Pittsburgh PA 15213, USA. 
  }
}

\begin{document}
\maketitle

\begin{strip}
  \centering
  \includegraphics[width=\textwidth]{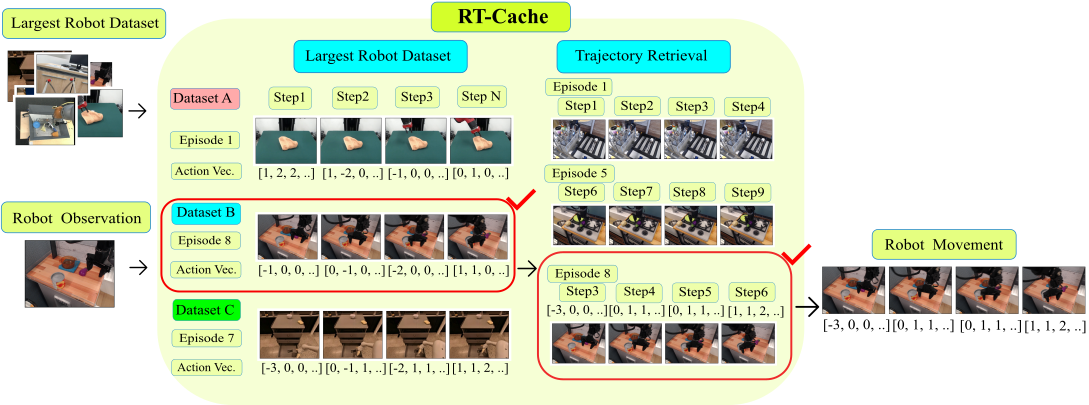}
    \captionof{figure}{
    \textbf{RT‑Cache at a glance.}
    \emph{Left:} We turn prior robot experience into an append‑only memory. Each episode contributes image–action snippets embedded in a shared vector space, independent of robot morphology.
    \emph{Center:} At deployment, the current observation is embedded and matched to similar snippets in memory; the highlighted segment is selected as the next controller state.
    \emph{Right:} The selected snippet is \emph{replayed} for the next $N$ steps, after which the system re‑queries—closing a retrieval loop. This training‑free “retrieval‑as‑control’’ view amortizes learning into stored experience, enabling real‑time, few‑shot adaptation by simply appending new episodes.
    }
    
  \label{fig:pipeline-overall}
  \vspace{-3mm}
\end{strip}

\thispagestyle{empty}
\pagestyle{empty}

\vspace{-2mm}
\begin{abstract}
Real robots are expected to repeat the same behavior in new environments with very little new data, yet modern controllers either incur heavy per‑step inference or require deployment‑time fine‑tuning. We propose \textbf{RT‑Cache}, a training‑free \emph{retrieval‑as‑control} pipeline that caches diverse image–action trajectories in a unified vector memory and, at test time, embeds the current frame to retrieve and \emph{replay} multi‑step snippets, replacing per‑step model calls. A hierarchical search keeps lookups sub‑second at million scale, shifting cost from compute to storage and enabling real‑time control on modest GPUs. Across real‑robot tasks and large open logs, RT‑Cache achieves higher success and lower completion time than strong retrieval baselines (approximately $2\times$ higher success and $\sim30\%$ faster in our settings), and a single‑episode anchoring study shows immediate adaptation to a more complex, contact‑rich task without fine‑tuning. RT‑Cache turns experience into an append‑only memory, offering a simple, scalable path to few‑shot deployment today and a foundation for multimodal keys and optional integration with high‑level policies. Project page: \url{https://rt-cache.github.io/}.

\end{abstract}

\begin{figure*}[t]
  \centering
  \includegraphics[width=\textwidth]{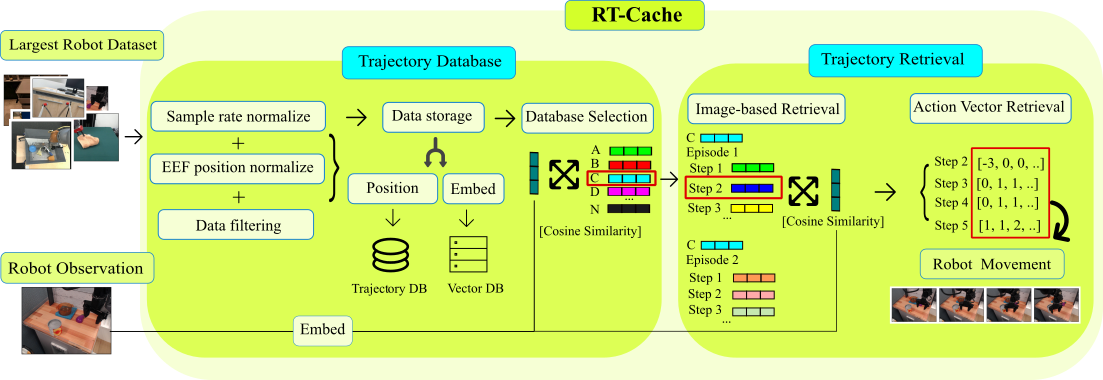}
    \caption{
    \textbf{RT‑Cache retrieval—from terabytes to real time.}
    \textbf{(1) Trajectory Database.} Heterogeneous logs are standardized (sample‑rate normalization, EEF Cartesian normalization, data filtering). Each image–action pair is stored in a \emph{Trajectory DB} and embedded into a \emph{Vector DB}.
    \textbf{(2) Trajectory Retrieval.} For each new observation, we embed the image and run a hierarchical search:
    dataset‑centroid filtering $\rightarrow$ sampled local index $\rightarrow$ cosine $k$‑NN over candidates. The selected snippet’s \emph{action vectors} are replayed for the next $N$ steps to command the robot, then the system re‑queries, forming a closed loop. This training‑free design replaces per‑step inference and keeps lookups sub‑second at million‑scale while adapting by simply appending new episodes to memory.
    }
    
  \label{fig:pipeline-detail}
\end{figure*}

\vspace{-1mm}

\section{INTRODUCTION}\label{sec:intro}
Robots are increasingly asked to repeat the \emph{same behavior} in \emph{new environments} with \emph{very little new data}. How can we deliver this in practice without heavy per‑task training or fine‑tuning, while keeping latency low enough for real hardware?

A natural candidate is the family of \emph{Vision–Language–Action} (VLA) models, which offer a unified interface for perception and control across embodiments \cite{kim2024openvla, black2024pi_0, brohan2023rt, team2024octo, kim2025fine}. Yet deployment remains costly: each control step often requires a large forward pass, inflating wall‑clock time and GPU demand; and adapting to unseen scenes typically requires task‑ or site‑specific fine‑tuning with non‑trivial data and compute. Even with speedups such as parallel decoding \cite{kim2025fine}, substantial computation at every step hampers real‑time use on modest hardware or multi‑camera humanoid setups.

A complementary direction is to \emph{reuse past experience} via retrieval \cite{pari2021surprising, du2023behavior, zha2024distilling, car2024plato, kwon2024toward, george2023one, stadie2017third, george2024vital, bartsch2024sculptdiff, young2021visual}. However, imitation‑style and policy‑retrieval methods typically assume a trained policy or adapters and focus on \emph{single‑step} selection on relatively small corpora; at deployment they often still require task‑specific fine‑tuning, and they rarely exploit \emph{multi‑step} reuse from large, heterogeneous datasets. As robot logs grow by orders of magnitude \cite{o2024open}, a system that scales retrieval to millions of states and \emph{replays} meaningful action segments—\emph{without} extra training—would bridge an important gap.

\textbf{Our RT-Cache system} reframes deployment as \emph{remember and replay}: cache successful real‑robot experience at scale and retrieve multi‑step snippets on the fly. RT‑Cache maintains a large \emph{Memory}, millions of image–action pairs, in a unified \emph{vector database} built from diverse trajectories. At runtime, the current camera frame is embedded into a shared space and compared against this memory; the top match (or an aggregate of the top-$K$) provides the next \(N\) actions. In effect, \emph{retrieval becomes the controller}, the robot executes the replayed sequence rather than generating each step with a large policy. Crucially, RT‑Cache is \emph{training‑free at deployment}: incorporating new scenes requires only adding their embeddings to the memory, no backpropagation, no per‑task fine‑tuning.

To keep lookups tractable on very large corpora, RT‑Cache uses a \emph{multi‑stage retrieval strategy}. We first narrow the search by dataset‑level centroids (or equivalent partitions), then query a small, sampled local index before a final $k$‑NN over candidates. This hierarchical filter turns exhaustive, minutes‑long global searches into sub‑second queries while preserving accuracy, \emph{enabling real‑time control}. Because the memory grows with use, even a few in‑domain examples can immediately \emph{anchor} retrieval in a new environment, enabling few‑shot adaptation without model updates. This \emph{retrieval‑as‑control} paradigm achieves low‑latency, practical deployment without any task‑specific optimization or encoder fine‑tuning.

\smallskip
\noindent
\textbf{Contributions.}
\begin{itemize}
  \item \textbf{RT-Cache}, a deployment‑time, training‑free pipeline that retrieves and \emph{replays multi‑step actions} from a large repository of real‑robot experience, effectively turning retrieval into the controller.
  \item A \textbf{scalable, multi‑stage retrieval} procedure (dataset‑centroid selection + small‑subset indexing + final $k$‑NN) that keeps lookup times sub‑second at million‑scale.
  \item A \textbf{unified trajectory memory} (vector database) spanning heterogeneous sources (Open‑X and additional real‑world logs) via standardized action representation and embeddings, enabling immediate few‑shot coverage by simply inserting new samples.
  \item Real‑robot experiments demonstrating that RT‑Cache \emph{cuts wall‑clock operation time} while maintaining or improving success; adding even a handful of in‑domain examples can flip zero‑shot failures into successes.
\end{itemize}

\section{RELATED WORKS}\label{sec:related}
\subsection{Vision–Language–Action (VLA) Models in Robotics}
Generalist VLA models have demonstrated strong cross‑task and cross‑embodiment generalization \cite{kim2024openvla, black2024pi_0, brohan2023rt, team2024octo, kim2025fine}, typically by pretraining on large, heterogeneous robot/vision corpora and then adapting to downstream settings. 
Despite this promise, three deployment pain points are common:
(1) \textbf{Per‑step compute}: each control step usually requires a large forward pass; even with engineering such as parallel decoding \cite{kim2025fine}, substantial computation occurs at every timestep, constraining control rates and increasing wall‑clock latency. 
(2) \textbf{Unseen‑environment fine‑tuning}: to reach reliable performance in new scenes (camera pose, lighting, workspace geometry), VLAs are often adapted with LoRA adapters or full fine‑tuning, which demands non‑trivial numbers of labeled steps and incurs additional iteration time. 
(3) \textbf{Resource footprint}: deployment frequently requires \emph{tens of GB} of GPU memory and tight I/O budgets, making continuous per‑step inference expensive on modest hardware or in multi‑camera humanoid settings. 
A further concern is \textbf{continual‑learning side effects}: repeated task‑specific updates can interfere with previously acquired skills \cite{kirkpatrick2017overcoming}. 
These factors have motivated complementary directions that lessen online compute and reduce (or eliminate) deployment‑time training while retaining broad task coverage.

\subsection{Retrieval‑Based Action Reuse and Planning}
Retrieval methods reuse prior experience by matching the current observation to stored trajectories and either copying actions or selecting data for a learned controller \cite{zha2024distilling,car2024plato,kwon2024toward,george2023one,stadie2017third,george2024vital,bartsch2024sculptdiff,young2021visual}. Image‑based approaches such as \textbf{VINN} perform per‑timestep RGB $k$‑NN and copy neighbor actions, yielding simple, training‑light deployment but suffering from single‑step drift and sensitivity to viewpoint/scale; most reports target modest index sizes rather than million‑scale, multi‑step replay \cite{pari2021surprising}. \textbf{Behavior Retrieval} instead uses retrieval to weight/select demonstrations for offline behavior cloning, leveraging large corpora but requiring task‑specific training/fine‑tuning and providing no test‑time retrieval, which can leave deployment mismatch uncorrected \cite{du2023behavior}. Language‑guided retrieval fetches demos or skills from text, typically assuming language‑conditioned policies and fine‑tuning after retrieval \cite{kuang2024ram}. Multimodal variants (vision with proprioception/depth/force) can disambiguate scenes but often mix retrieval with online learning or controller adaptation, making it hard to isolate retrieval‑only gains and rarely reporting sub‑second search at million scale \cite{memmel2024strap,guo2025srsa}. In summary, most prior work uses retrieval for a controller or for per‑step copying; training‑free, \emph{multi‑step} replay from a large, heterogeneous memory with sub‑second lookup remains under‑explored.

In contrast, \emph{RT‑Cache} treats retrieval \emph{as the controller}: a training‑free system that replays multi‑step snippets from a million‑scale, heterogeneous memory via a hierarchical search with sub‑second latency. We deliberately start with \emph{image keys} because RGB is the only modality shared across robot morphologies and it stresses the highest‑bandwidth index, enabling a clean, reproducible comparison to image‑based baselines such as VINN and Behavior Retrieval \cite{pari2021surprising,du2023behavior}. Execution still operates in a unified cartesian action space, while retrieval remains modality‑agnostic: proprioception, depth, or language cues can be appended to the index without changing the control loop. This separates the core claim—training‑free, multi‑step replay at scale—from key‑design choices, and leaves a full multimodal benchmark to future work.

\begin{figure}[t]
    \centering
    \includegraphics[width=0.92\linewidth]{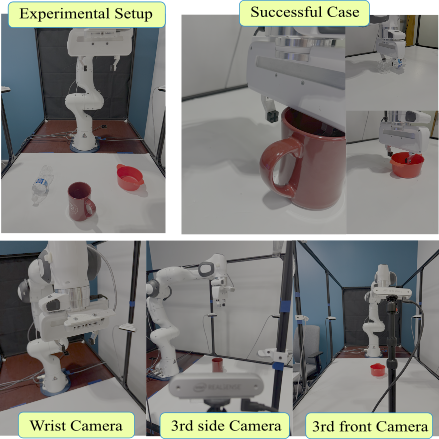}
    \caption{
    \textbf{Multi‑camera test arena and success criterion.}
    A 7‑DoF Franka Emika Panda manipulator operates on a tabletop observed by three Intel RealSense D415 cameras: \textit{wrist}, \textit{third‑person side}, and \textit{third‑person front}. One of three objects (bottle, mug, bowl) is placed before each rollout. A trial is \emph{successful} when the end‑effector reaches the object’s graspable region by the final step. Top‑left: arena; top‑right: example success frames; bottom: the three viewpoints.
    }
    \label{fig:cam-setup}
    \vspace{-3mm}
\end{figure}
\section{METHODS}\label{sec:methods}
This section details our approach for real-time robot control. 
We first describe how we build a large-scale, unified dataset from the diverse trajectories in the Open-X Embodiment collection \cite{o2024open}. 
We then present the \emph{RT-Cache Pipeline}, a retrieval-based framework that leverages this unified dataset to accelerate multi-step action selection in real-world environments.

\vspace{-1mm}
\subsection{Building a Unified Dataset Collection}\label{subsec:openx}
\subsubsection{Open-X Embodiment}\label{subsec:openx}
Our work leverages the Open-X Embodiment dataset, which includes over one million
real-robot trajectories from 22 platforms (single-arm to bi-manual and
quadrupeds). Contributed by 34 labs worldwide, these 60 sub-datasets use RLDS
format \cite{kim2024openvla} and span diverse action spaces (Cartesian vs.\ joint)
and sensor modalities (RGB, depth, point clouds).

\subsubsection{Dataset Processing}\label{subsec:openx-procesing}
The Open-X dataset spans control frequencies from 3\,Hz to 30\,Hz and includes actions expressed in position, velocity, or joint-space commands. To ensure uniformity, we adopt a three-step \emph{unification} procedure:
\begin{enumerate}
  \item \textbf{Sampling rate standardization:} Convert all trajectories to 10\,Hz. Lower-frequency data are linearly interpolated, while higher-frequency data are subsampled.
  \item \textbf{EEF position representation:} Represent every action as an end-effector (EEF) position in Cartesian space. For velocity-based datasets (e.g., 0.2\,m/s), integrate each velocity over 0.1\,s to approximate positional displacement.
  \item \textbf{Filtering by action shape:} Exclude any episodes lacking a 7-D action shape or those using only joint-space commands.
\end{enumerate}

This standardization process yields a consistent, 7-D Cartesian action representation across all sub-datasets, simplifying subsequent retrieval tasks.

\subsubsection{Database Setup}\label{subsec:db-setup}
We adopt a \textbf{two-tiered database strategy} to efficiently store raw robot data alongside high-dimensional embeddings:

\paragraph{MongoDB for Raw Data.}
We use MongoDB to store time-step information (e.g., timestamps, unified 7-D actions, RGB frames). Each entry corresponds to a single step, indexed by \texttt{(episode\_id, step\_id)} for rapid lookups and flexible filtering.

\paragraph{Vector Database for Embeddings.}
For similarity-based retrieval, we maintain a separate \emph{vector database} (Qdrant). We compute and store one embedding per state:
\begin{itemize}
  \item \textbf{DINOv2 \cite{oquab2023dinov2} image features} (1024-dimensional)
  \item \textbf{SigLIP \cite{zhai2023sigmoid} image features} (1152-dimensional)
\end{itemize}
We concatenate these features into a single 2176-dimensional vector, which empirically outperforms either encoder alone. At query time, the incoming camera frame is embedded similarly, and a top-$K$ nearest-neighbor search (e.g., via cosine distance) retrieves the most relevant states.

The vector database returns candidate state IDs, which map back to MongoDB records for the corresponding raw data. This separation between \emph{similarity lookup} (in the vector DB) and \emph{record storage} (in MongoDB) allows our system to scale to millions of states without compromising retrieval speed.

\subsection{RT-Cache Pipeline}\label{subsec:pipeline}
\subsubsection{Overall Retrieval-Based Pipeline}\label{subsec:ret-pipeline}
Our system converts each new camera image into a high-dimensional embedding and uses it to query a large database of prior robot experiences for candidate actions. The retrieval pipeline proceeds in four main stages, each with defined parameters:

\paragraph{Stage 1: Embedding the Current Observation}


\begin{itemize}
    \item Let $I_t$ be the current RGB frame at time~$t$.
    \item We feed $I_t$ through two pretrained encoders, DINOv2 and SigLIP, yielding feature vectors $\mathbf{d}_t \in \mathbb{R}^{1024}$ and $\mathbf{s}_t \in \mathbb{R}^{1152}$, respectively.
    \item We then concatenate and $\ell_2$-normalize these vectors to form a single \textbf{target embedding}:
    \begin{equation}
        \mathbf{e}_t = \mathrm{Norm}\Bigl([\mathbf{d}_t \,\|\, \mathbf{s}_t]\Bigr),
        \label{eq:target-embed}
    \end{equation}
    where $[\cdot \,\|\, \cdot]$ denotes vector concatenation and $\mathrm{Norm}(\cdot)$ denotes $\ell_2$-normalization.
\end{itemize}

\paragraph{Stage 2: Database Selection}
A naive search over billions of embeddings is computationally infeasible, so we employ two  different filtering steps that drastically narrow down the candidate space before a final $k$-nearest-neighbor query:

\begin{enumerate}
    \item \textbf{Dataset-Centroid Stage.}  
    We partition the entire Open-X corpus by dataset (e.g., each lab’s subset). For dataset~$d$, we precompute a centroid $\mathbf{c}_d$ by averaging all embeddings within that dataset:
    \begin{equation}
        \mathbf{c}_d = \frac{1}{N_d} \sum_{i=1}^{N_d} \mathbf{f}_{d,i}, 
        \quad d = 1,2,\ldots,D,
        \label{eq:centroids}
    \end{equation}
    where $\mathbf{f}_{d,i}\in \mathbb{R}^{2176}$ is the $i$-th embedding and $N_d$ is the total number of embeddings in dataset $d$.  
    At query time, we compute the distance (e.g., cosine distance) between the target embedding $\mathbf{e}_t$ and each centroid $\mathbf{c}_d$, then select the top $m$ datasets whose centroids are closest:
    \begin{equation}
        d^*_{1:m} = \arg \min_{d \in \{1,\dots,D\}} \text{dist}(\mathbf{e}_t, \mathbf{c}_d).
    \end{equation}

    \item \textbf{Small Subset Stage.}  
    For each of these $m$ shortlisted datasets, we \emph{randomly sample} (or \emph{cluster-sample}) a small subset of size $S$ (e.g., a few thousand) to form a local index. We then perform a finer-grained $k$-nearest-neighbor ($k$-NN) search on each local index using $\mathbf{e}_t$, retrieving the top candidates from each.
\end{enumerate}
This hierarchical filtering limits retrieval to $\mathcal{O}(m \times S)$ embeddings rather than $\mathcal{O}\!\bigl(\sum_d N_d\bigr)$, greatly speeding up queries.

\paragraph{Stage 3: Similarity Computation and Candidate Selection}
After pooling the candidates from all local indices, we calculate \emph{cosine similarity} between the target embedding $\mathbf{e}_t$ and each candidate embedding $\mathbf{f}_j$:
\begin{equation}
    \text{sim}(\mathbf{e}_t, \mathbf{f}_j) 
    \;=\; \frac{\mathbf{e}_t \cdot \mathbf{f}_j}{\|\mathbf{e}_t\|\;\|\mathbf{f}_j\|}.
    \label{eq:cosine-sim}
\end{equation}
We rank the candidates by similarity and select the top $K$ (e.g., $K=50$). Let $\{\mathbf{f}_{j}\}_{j=1}^K$ denote these final neighbors.

\paragraph{Stage 4: Retrieval and Action Execution}
Each embedding $\mathbf{f}_{j}$ maps to a database record containing a \emph{trajectory snippet}, i.e., the next $N$ actions $\mathbf{a}_{t:t+N}^{(j)}$. We consider two approaches:
\begin{itemize}
    \item \textbf{Single Best Neighbor:} Use only the highest-scoring neighbor $j^*$ and retrieve $\mathbf{a}_{t:t+N}^{(j^*)}$, executing these $N$ actions with no further inference calls.
    \item \textbf{Averaged Actions (Multi-Neighbor):} Compute an average of actions from the top $K$ neighbors:
    \begin{equation}
        \overline{\mathbf{a}}_{t:t+N} = \frac{1}{K}\sum_{j=1}^K \mathbf{a}_{t:t+N}^{(j)},
        \label{eq:avg-actions}
    \end{equation}
    which can reduce noise if multiple neighbors are closely related.
\end{itemize}

\begin{figure*}[t]      
  \centering
  \includegraphics[%
      width=\textwidth,            
      height=.5\textheight,      
      keepaspectratio              
]{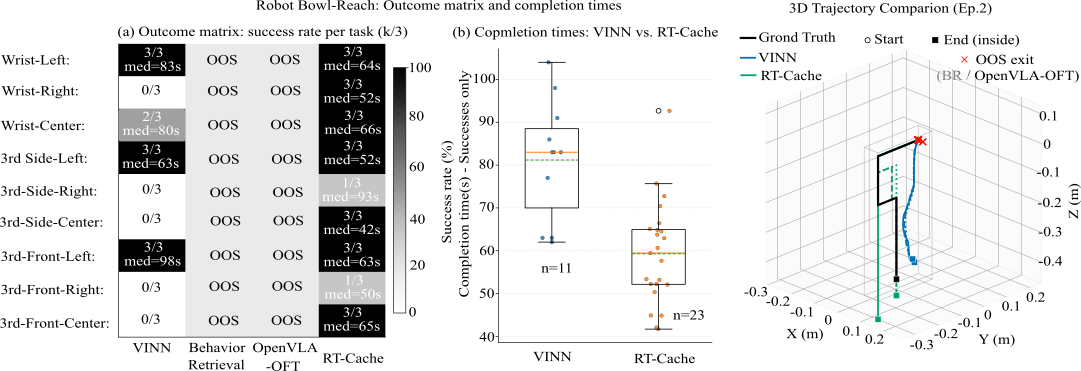}
    \caption{
    \textbf{Robot bowl‑reach: outcomes, time, and path quality.}
    \textbf{(a)} Outcome matrix reporting \emph{success rate per task} (\(k/3\) trials) across camera views/placements and methods. Cell shading encodes success percentage; overlaid text shows \(k/3\) and the median completion time (s) for \emph{successes only}.
    \textbf{(b)} Completion times for successful runs. Boxes show the interquartile range (IQR) and median; points are individual trials; \(n\) beneath each box is the number of successes. RT‑Cache exhibits a lower median and tighter spread than VINN.
    \textbf{(c)} 3D end‑effector trajectories for the same episode (legend as labeled). RT‑Cache closely tracks ground truth, whereas VINN drifts. Behavior‑Retrieval and OpenVLA‑OFT terminate out-of-workspace (OOS) in this episode and are omitted for clarity.
    }
  \label{fig:offline-test}
\end{figure*}

\subsection{Experimental Setup}\label{subsec:evaluation}

We evaluate RT-Cache on a simplified manipulation task where a 7-DoF \emph{Franka Emika Panda} robot moves its end-effector (EEF) toward a specific object on a tabletop. Three Intel RealSense D415 cameras provide complementary views: a lateral \textbf{Side Cam}, a wide-angle \textbf{Front Cam}, and a \textbf{Wrist Cam} mounted on the robot arm (Fig.~\ref{fig:cam-setup}). We randomly place three objects—a plastic bottle, a ceramic mug, and a plastic bowl—at varied positions. A trial is deemed \emph{Successful} if the EEF is within the object’s graspable region by the final step.

\paragraph{Scenarios}
We test two setups to gauge RT-Cache’s adaptability:
\begin{itemize}
    \item \textbf{Zero-Shot}: No in-domain data exist for the specific object/camera/pose combination; we ask whether purely visual similarity can retrieve a relevant multi-step snippet.
    \item \textbf{Few-Shot}: We add up to 27 short demonstration episodes (each $\sim$15 steps) for the new scene to assess how minimal in-domain examples anchor retrieval.
\end{itemize}

\paragraph{Data Collection}
We recorded short trajectories involving 2–3 objects placed at left/right/front/back table locations, logging at each step: A \textbf{3D Cartesian} action (EEF translation), and a \textbf{single} wrist or third-person camera image.

\paragraph{Baselines and Implementation Details}
For comparability, we include two retrieval-style baselines and our method, using standard configurations:
\begin{itemize}
    \item \textbf{VINN}~\cite{pari2021surprising}: Train a BYOL encoder (ResNet‑50) on demonstration images to obtain 2048‑D features; at deployment, encode the query frame and perform cosine $k$‑NN over stored demo frames. The action is a similarity‑weighted average of the $k$ neighbors’ actions. No policy is executed at test time; adaptation is achieved by inserting new demos into the index.
    \item \textbf{Behavior Retrieval}~\cite{du2023behavior}: Fit a $\beta$‑VAE over state–action pairs to learn a behavior similarity space; use it to reweight a behavior cloning (BC) policy via retrieved demonstrations from a larger unlabeled set. We samples 25\% (~25,000 samples) from a large unlabeled Open-X dataset (100,000+ samples) to weight the behavior cloning policy training alongside. At deployment, execute the trained BC policy (no retrieval in the loop). Optional interactive corrections (e.g., HG‑DAgger) are not used in main results.
    \item \textbf{RT‑Cache (Ours)}: Use frozen foundation features (DINOv2 + SigLIP; concatenated 2176‑D) stored in a Qdrant vector database; neighbor IDs map to MongoDB to fetch actions. At runtime we retrieve top‑$K$ matches and replay $N$‑step snippets (or a similarity‑weighted aggregate) with \emph{no training or fine‑tuning at deployment}. Hierarchical filtering settings are as described in Sec.~\ref{subsec:pipeline}.
\end{itemize}

\begin{table}[t]
  \centering
  \footnotesize
  \renewcommand{\arraystretch}{1.15}
  \setlength{\tabcolsep}{5pt}
  \caption{\textbf{Compute and storage budget across methods.}
  GPU and disk ranges reflect typical runs in our setup.
  \emph{Gradient updates?} is \checkmark{} when any gradient-based parameter updates
  are performed in our pipeline (e.g., encoder fine-tuning, VAE/policy training,
  parameter-efficient fine-tuning); \(\times\) indicates a fully forward-only pipeline.
  RT-Cache uses frozen encoders and performs no gradient updates.}
  \label{tab:system_usage}

  \begin{tabularx}{\linewidth}{@{}l
      >{\raggedleft\arraybackslash}X
      >{\raggedleft\arraybackslash}X
      c@{}}
    \toprule
    \textbf{Configuration} & \textbf{GPU mem. (GB)} & \textbf{Ext. disk (GB)} & \textbf{Gradient updates} \\
    \midrule
    \textbf{RT-Cache (Ours)} &
      \SIrange{8}{10}{} &
      \(\sim\)\num{100}\footnotemark[1] &
      $\times$ \\
    \textbf{Behavior Retrieval \cite{du2023behavior}} &
      \SIrange{4}{6}{} &
      $<\,\num{1}$ &
      $\checkmark$ \\
    \textbf{VINN \cite{pari2021surprising}} &
      \SIrange{6}{8}{} &
      $<\,\num{1}$ &
      $\checkmark$ \\
    \textbf{OpenVLA-OFT \cite{kim2025fine}} &
      \SIrange{16}{24}{} &
      \num{15}\,(per ckpt.) &
      $\checkmark$ \\
    \bottomrule
  \end{tabularx}

  \vspace{2pt}
  \footnotesize
  \emph{Notes.} RT-Cache uses \emph{frozen} foundation features; we compute embeddings and build an ANN index only (no gradient updates).\\
  \footnotetext[1]{All values in gigabytes. Breakdown: \(\sim\)95 for image embeddings + \(\sim\)3.1 for metadata (MongoDB).}
\end{table}

\begin{figure*}[t]      
  \centering
  \includegraphics[%
      width=\textwidth,            
      height=.5\textheight,      
      keepaspectratio              
]{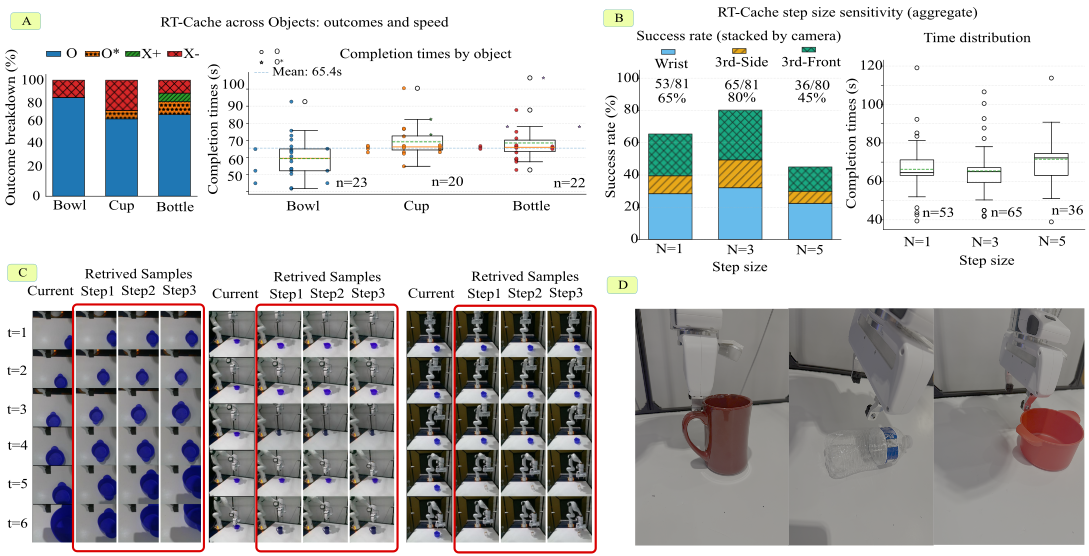}
    \caption{\textbf{Ablations and qualitative analysis.}
    (A) Object robustness across \emph{bowl/cup/bottle}. Stacked bars encode outcomes:
    O (success), O* (success with path adaptation), X- (near miss),
    X+ (wrong turn); boxplots show completion times for \emph{successes only}.
    (B) Horizon sensitivity vs.\ snippet length \(N\!\in\!\{1,3,5\}\): left—overall success
    rate aggregated over \emph{wrist}, \emph{3rd‑side}, \emph{3rd‑front}; right—completion‑time
    distributions (\emph{successes only}). (C) Viewpoint‑aligned replay. For each camera
    (wrist, 3rd‑side, 3rd‑front), columns show \emph{Current} and replayed frames
    \emph{Step\,1–Step\,3}; rows are independent trials. Retrieval stays within the active
    view and advances coherently over \(N\) steps. (D) Typical failure cases.}
  \label{fig:offline-test}
\end{figure*}

\paragraph{Fine-Tuning Details (VLA Baseline)}
For the VLA baseline, we start from a pretrained OpenVLA‑OFT checkpoint and apply Low‑Rank Adaptation (LoRA) with: rank $32$, learning rate $5\times10^{-4}$ (decayed after $100$k steps), and a maximum of $150{,}000$ steps using a single third‑person viewpoint. 

\section{RESULTS AND DISCUSSION}\label{sec:results-discussion}
We compare \emph{RT‑Cache} with VINN, Behavior Retrieval (BR), and OpenVLA‑OFT. We first report outcome rates and completion times, then analyze trajectory quality.

\subsection{Performance Comparison}\label{subsec:perf}

\paragraph{Reporting protocol (zero‑shot vs.\ few‑shot)}
We first evaluated a strict zero‑shot setting (no in‑domain frames for a given object–camera–pose combination) and observed \textbf{0\% success across all methods and views}. Consequently, all quantitative results below are reported in the \emph{few‑shot} regime, where we insert up to 27 short in‑domain episodes to anchor retrieval. This reflects the intended deployment mode and highlights the benefits of multi‑step replay under minimal supervision.

\paragraph{Success rate and completion time}
Across all tasks, \emph{RT‑Cache} achieves higher success with faster completions than VINN (Fig.~4a–b). VINN’s coverage is narrow (mainly wrist–center), whereas RT‑Cache succeeds in third‑person views where VINN fails. Aggregated over \emph{successes only}, RT‑Cache shows a lower median time and tighter spread ($\approx$30\% faster than VINN). The speedup stems from replaying multi‑step snippets, which reduces corrective micro‑steps. (Per‑task medians appear in the figure caption.)

\paragraph{Trajectory quality}
Overlaid 3D paths (Fig.~4c) show RT‑Cache closely tracking the ground‑truth trajectory, while VINN exhibits lateral drift and late oscillations. This drift is consistent with per‑timestep, similarity‑weighted averaging that accumulates small biases. 

\paragraph{Other baselines}

In this setup, \emph{Behavior Retrieval (BR)} and \emph{OpenVLA-OFT} frequently triggered an \emph{out-of-workspace (OOS)} stop, i.e., the end-effector left the allowed workspace/camera bounds, and are counted as failures (Fig.~4c). BR executes a policy trained offline and is sensitive to cross-domain calibration; OpenVLA-OFT still pays per-step inference cost and struggled under limited in-domain data. We include these results for completeness.


\begin{table}[h]
\centering
\caption{
\textbf{Operation times for different database-selection strategies.}
A naive full-database query can exceed 300\,s, making real-time usage infeasible.
In contrast, our sampling both complete in about 0.1\,s on the same corpus, enabling real-time retrieval.
}
\label{tab:op_time}
\begin{tabular}{l l}
\toprule
\textbf{Selection Method} & \textbf{Operation Time} \\
\midrule
Full-DB Search & Time-out ($>$300\,s) \\
Dataset-Centroid Stage & $\sim$0.1\,s \\
Small Subset Stage & $\sim$0.1\,s \\
\bottomrule
\end{tabular}
\end{table}

\begin{figure*}[t]      
  \centering
  \includegraphics[%
      width=\textwidth,            
      height=.9\textheight,      
      keepaspectratio              
]{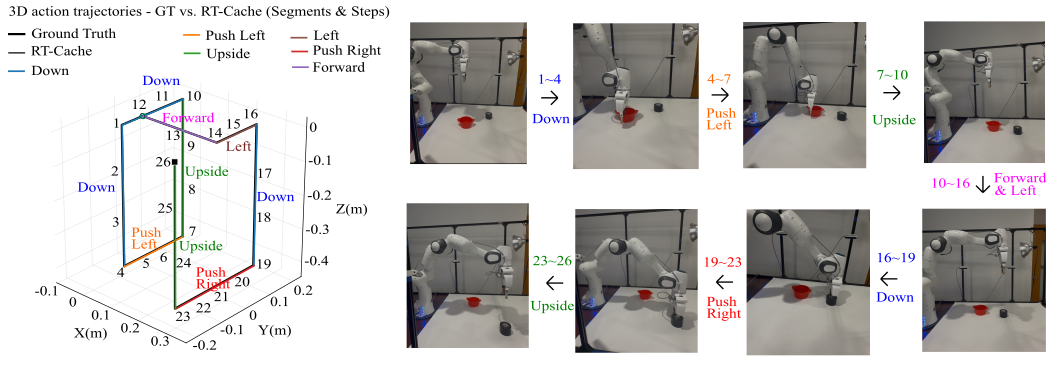}
    \caption{\textbf{Single‑episode anchoring on a pushing task.}
    \emph{Left:} 3D end‑effector trajectories for ground truth vs.\ RT‑Cache. Colors denote motion segments; RT‑Cache closely tracks the reference path.
    \emph{Right:} Third‑person snapshots with colored arrows indicating end‑effector motion. The memory initially contained no in‑domain data; adding \emph{one} short episode (no training/fine‑tuning) created an anchor that RT‑Cache retrieves and replays, yielding a trajectory that qualitatively coincides with the ground truth. This illustrates \emph{retrieval‑as‑control}: a single exemplar can unlock real‑time execution.}
  \label{fig:offline-test}
\end{figure*}

\subsection{System efficiency and compute budget}\label{subsec:results-system}
Table~\ref{tab:system_usage} compares GPU/disk budgets and fine‑tuning needs. \emph{RT‑Cache} runs in \(\sim\)8–10\,GB GPU, shifts cost to storage (\(\sim\)100\,GB embeddings+metadata in a vector DB), and requires \emph{no} deployment‑time training or fine‑tuning. VINN and Behavior Retrieval \emph{train} their own encoders/policies (BYOL; VAE+BC) with smaller footprints, while OpenVLA‑OFT is heaviest (16–24\,GB GPU) and adds per‑checkpoint disk from fine‑tuning. Net effect: RT‑Cache trades SSD space for compute, amortizing model work into an append‑only memory that enables real‑time control on modest GPUs.

\subsection{Ablation Studies (Fig. 5)}\label{subsec:ablations}

\paragraph{Object robustness}
With a few in‑domain episodes, RT‑Cache is reliable across objects and views. Overall success is \(96\%\) with the wrist‑cam setup (median time \(66\,\mathrm{s}\)) and \(93\%\) from the third‑person front view (\(63\,\mathrm{s}\)), while the third‑person side view drops to \(52\%\) (\(77\,\mathrm{s}\)). Per‑object totals span \(78\text{--}100\%\) (bowl), \(22\text{--}100\%\) (cup), and \(56\text{--}100\%\) (bottle), with completion times between \(60\text{--}68\,\mathrm{s}\) in the easier views. In contrast, the zero‑shot setting yields \(0\%\) success across objects. Most errors are near‑misses rather than wrong‑turn failures (Fig.~5A).

\paragraph{Horizon sensitivity}
In Fig. 5.B, varying the snippet length \(N\in\{1,3,5\}\) trades reactivity for speed: \(N{=}1\) adapts every step but adds lookup overhead, \(N{=}5\) is fastest when the match is perfect yet less forgiving to small mismatches, and \(N{=}3\) consistently gives the best speed–accuracy compromise. Aggregate success/time values are in the caption. Unlike VINN (per‑timestep copying) or Behavior Retrieval (offline policy, no test‑time retrieval), RT‑Cache exploits \(N{>}1\) \emph{multi‑step replay} without any training or fine‑tuning.

\paragraph{How similar is the retrieved path to the original in (Fig. 5.C)?}
\begin{itemize}
  \item \textbf{1 — Object-Centric retrieval.} Each retrieved snippet (e.g., wrist view) targets mostly same object and viewpoint but shows small variations—akin to a human repeating a skill with minor differences.
  \item \textbf{2 — Temporal-Coherence retrieval.} The bottom sequences illustrate progression from the “current” frame through subsequent steps, showing how RT‑Cache continually selects forward‑looking frames consistent with the active view.
  \item \textbf{3 — Viewpoint‑specific retrieval.} Queries respect the camera: wrist‑cam retrieves wrist‑cam snippets; side‑cam retrieves side‑cam, and so on. This viewpoint consistency noticeably improves retrieval accuracy in our trials.
\end{itemize}

\paragraph{Retrieval latency and throughput}\label{subsec:results-speed}
Table~\ref{tab:op_time} compares retrieval times for different database-selection strategies. An exhaustive nearest-neighbor search over millions of embeddings can take minutes, effectively making real-time tasks impossible. In practice, traditional systems might time out because they are not designed for massive datasets.  Instead, our strategy completes lookups in under one second, even on large corpora, keeping RT-cache practical for real-world deployments where every second matters—such as human–robot collaborative settings.

\paragraph{Failure modes and mitigations}
Figure 5.D illustrates typical errors, such as minor positional offsets or incorrect gripper orientation. These near-misses often occur when an object is partially occluded or the snippet comes from a slightly different pose. 

\paragraph{Embedding keys (CLIP \cite{radford2021learning} vs.\ DINOv2+SIGLIP \cite{oquab2023dinov2, zhai2023sigmoid})}
In main experiments we use DINOv2+SigLIP concatenation to improve recall. Using image features as the retrieval key, DINOv2+SigLIP achieves roughly \emph{2×} higher success than CLIP, especially from the wrist view, while CLIP struggles in third‑person‑front scenes.

\subsection{Advanced task with single‑episode anchoring (Fig.~6)}\label{subsec:advanced}
\textbf{Setup.} We evaluate a more \emph{complex, contact‑rich} tabletop pushing task from a third‑person view. With \emph{no} initial in‑domain data, we insert \emph{one} short episode by upserting its frame embeddings into the vector database and logging cartesian action deltas for replay.

\textbf{Result.} At test time, RT‑Cache matches the scene to that single exemplar and replays the \(N\)-step snippet end‑to‑end, \emph{without} fine‑tuning or gradients, producing a trajectory that qualitatively coincides with ground truth. A single example is enough to run the behavior in real time.



\section{LIMITATIONS AND FUTURE DIRECTIONS}\label{sec:limitations}
RT‑Cache retrieves with \emph{image‑only} keys. This keeps deployment simple and universal, but can yield near‑misses under occlusion, fine pose changes, or heavy clutter/dynamics. The controller also replays short \(N\)-step segments rather than composing longer, multi‑stage skills, and sustained growth of the memory raises storage/indexing efficiency questions.

We will broaden evaluation to multi‑object, cluttered and dynamic scenes, and report metrics that isolate retrieval quality from execution \cite{bartsch2025pinchbot, schaldenbrand2024cofrida, schaldenbrand2022frida}. On the modeling side, we plan to add multimodal keys (proprioception/depth/language) with confidence‑aware \cite{kuang2024ram, guo2025srsa}, dynamic horizons \cite{memmel2024strap}, and to pair retrieval‑first control with a lightweight VLA fallback or memory‑authoring policy. We will explore compression to keep the append‑only memory fast at scale.

\vspace{1mm}

\section{CONCLUSIONS}\label{sec:conclusions}
We introduced \emph{RT‑Cache}, a training‑free \emph{retrieval‑as‑control} pipeline that replays multi‑step snippets from a large vector memory; a hierarchical search keeps million‑scale lookups sub‑second, shifting cost from compute to storage and enabling real‑time control on modest GPUs. In tabletop reaching, RT‑Cache achieves higher success and $\sim$30\% lower median completion time than VINN while avoiding cross‑domain failures seen with Behavior‑Retrieval–style policies; a moderate horizon ($N{=}3$) balances speed and reactivity, and viewpoint‑aligned replay reduces drift. Using DINOv2+SigLIP keys improves recall at negligible runtime cost, and a single‑episode anchoring study shows immediate adaptation on a contact‑rich task with no fine‑tuning. The framework is readily extensible to multimodal keys and can interoperate with VLA policies as a fallback or for memory authoring.




\bibliographystyle{IEEEtran}  
\bibliography{references}

\end{document}